\documentclass[11pt]{article}

\usepackage[final]{acl}

\usepackage{times}
\usepackage{latexsym}

\usepackage[T1]{fontenc}

\usepackage[utf8]{inputenc}

\usepackage{microtype}

\usepackage{inconsolata}

\usepackage{graphicx}

\usepackage{algorithm}
\usepackage{algpseudocode}

\usepackage{booktabs}
\usepackage{multirow}
\usepackage{graphicx}
\usepackage{xcolor}

\usepackage{booktabs}
\usepackage{multirow}
\usepackage[table]{xcolor}
\usepackage{array}

\usepackage{array}
\newcolumntype{C}[1]{>{\centering\arraybackslash}p{#1}}

\newcolumntype{L}[1]{>{\raggedright\arraybackslash}p{#1}}
\newcolumntype{M}[1]{>{\centering\arraybackslash}m{#1}}

\usepackage{amsmath}
\usepackage{amssymb}

\title{Unifying Detection and Adaptation in Task-Free Continual Learning}

\author{ 
  \textbf{Dezheng Han}\textsuperscript{1}, 
  \textbf{Anbang Zhang}\textsuperscript{2}, 
  \textbf{Zhihao Zhu}\textsuperscript{3,4}, 
  \textbf{Shuaishuai Guo}\textsuperscript{1,*}\\
  \textsuperscript{1}School of Control Science and Engineering, Shandong University, Jinan, China \\
  \textsuperscript{2}Department of ECE, The Hong Kong University of Science and Technology, Hong Kong \\
  \textsuperscript{3}National Medicine-Engineering Interdisciplinary Industry-Education Integration Innovation Platform, \\
  Shandong University, Jinan, China \\
  \textsuperscript{4}Shandong Key Laboratory: Magnetic Field-free Medicine \& Functional Imaging, \\
  Qilu Hospital of Shandong University, Jinan, China \\
  \{dezhenghan, 202415670\}@mail.sdu.edu.cn, 
  azhangay@connect.ust.hk, 
  shuaishuai\_guo@sdu.edu.cn
}

\begin{document}
\maketitle

\begingroup
\renewcommand{\thefootnote}{}
\footnotetext{Code is available at: \url{https://github.com/SSG2019/FiUni}.}
\addtocounter{footnote}{-1}
\endgroup
\begin{abstract}
To mitigate catastrophic forgetting in downstream continual learning (CL) for large language models (LLMs), existing methods typically constrain parameter updates or introduce task-specific adaptation modules. However, these methods often rely on explicit task boundaries during training, limiting their applicability to realistic task-free scenarios. In this paper, we propose a \textbf{Fi}sher-guided \textbf{uni}fied (\textbf{FiUni}) framework for batch-level task detection and parameter-efficient continual adaptation.
FiUni is motivated by a key observation about the Fisher information matrix (FIM) of pre-trained models: the orthogonality among the principal subspaces of its Kronecker-Factored Approximate Curvature (K-FAC) approximation, estimated from a small number of downstream task samples, can reflect the similarity between different tasks.
Based on this observation, FiUni constructs FIM-derived frozen subspaces to guide low-rank adaptation (LoRA), while matching the Fisher principal subspace of each incoming batch window with historical subspaces. This enables FiUni to adaptively determine whether to reuse existing knowledge, expand a related subspace, or create a new subspace, dynamically balancing knowledge sharing and task isolation. Experiments show that FiUni can effectively infer latent batch-level task affiliations and achieve competitive performance against advanced task-aware CL methods with fewer trainable parameters.
\end{abstract}

\begin{figure*}[t]
    \centering
    \includegraphics[width=\textwidth]{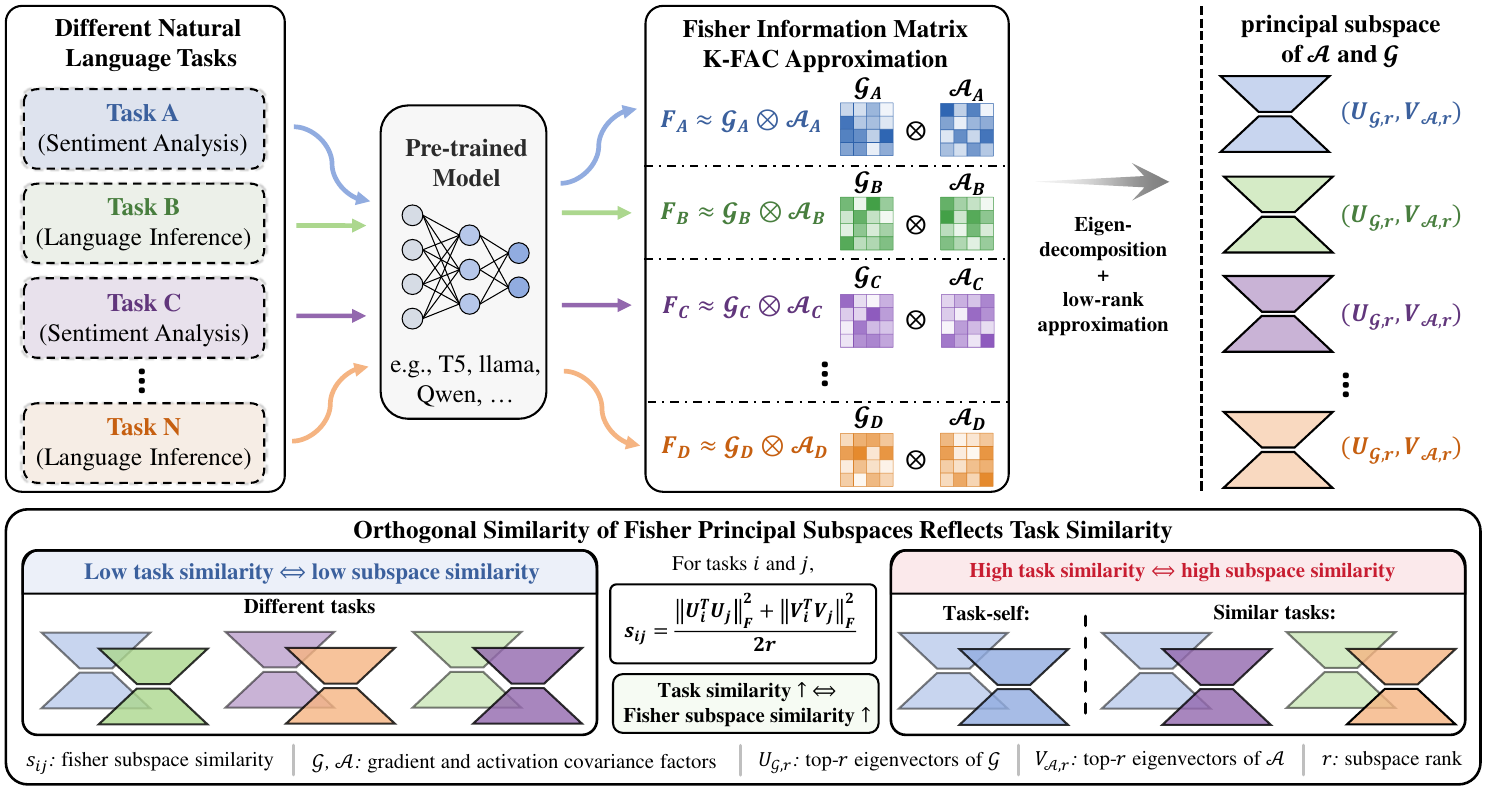}
    \caption{Fisher principal subspaces encode task similarity. For each downstream task, we estimate the K-FAC approximation and extract the top-$r$ principal subspaces of its gradient and activation covariance factors. Higher subspace similarity indicates more similar task geometry, while lower similarity indicates stronger orthogonality.
    }
    \label{fig1}
\end{figure*}

\section{Introduction}

With the remarkable generalization capabilities of large language models (LLMs) \cite{t5, llama, qwen} across diverse downstream tasks, enabling LLMs to continuously adapt to dynamic environments has emerged as a critical research problem \cite{cl_survey3}. However, direct full-parameter fine-tuning for continual learning (CL) \cite{cl_survey1, cl_survey2} suffers from prohibitive computational and storage costs, and inevitably leads to severe catastrophic forgetting \cite{cf2}, where the model loses performance on old tasks after learning new ones. Parameter-efficient fine-tuning (PEFT) \cite{peft_survey1,peft_survey2}, which updates only a small number of original or additional parameters, provides a more feasible solution to this problem. Following this idea, existing parameter-efficient continual learning methods typically manage the low-dimensional parameter update spaces of different tasks, thereby reducing inter-task interference while maintaining fine-tuning efficiency \cite{o_lora, sparta, ella}.


Despite promising performance in multi-task continual learning, existing methods typically rely on task-aware assumptions, where task boundaries or IDs are available for allocating and switching parameter subspaces. In contrast, real-world continual learning is often task-free, with data arriving in an online stream without explicit boundary annotations or task identities \cite{task_free_cl}. Thus, the model must autonomously infer latent task structure while adapting its parameters to incoming data. Although recent methods exploit model-internal task-related signals to relax this assumption \cite{migu}, they mainly focus on task-label-free forgetting mitigation, rather than explicitly unifying task identification with the isolation and sharing of PEFT subspaces.


To address these issues, we need a meaningful geometric signal that can characterize the relationships among different incoming data and further guide parameter-efficient adaptation.
Fisher geometry \cite{natural-gradient} provides a promising perspective for this purpose.
Recent studies have shown that the  Kronecker-Factored Approximate Curvature (K-FAC) \cite{k-fac} of the Fisher information matrix (FIM), estimated from a small number of downstream samples of different tasks on a pre-trained model, can serve as effective fixed low-rank update bases to guide LoRA along parameter-sensitive directions \cite{filora}.
This motivates us to further explore whether Fisher principal subspaces can be used to model latent task relationships among different downstream data.

Under this perspective, we discover that in the K-FAC \cite{k-fac} approximation of the FIM estimated on a few downstream samples, the orthogonality of the principal subspaces, spanned by the top-$r$ eigenvectors of the gradient and activation covariance factors, naturally reflects the geometric similarity between distinct tasks, as shown in Figure~\ref{fig1}. Specifically, data from identical or related tasks tend to exhibit higher Fisher principal subspace similarity, whereas those from unrelated tasks tend towards orthogonality. This indicates that for pre-trained models, the Fisher principal subspace not only characterizes the sensitive update directions for the current data but also serves as a geometric signal to identify the latent task affiliation of each batch in an online data stream. Here, a latent task does not necessarily align perfectly with human-annotated discrete tasks, but rather represents an implicit learning phase perceived by the pre-trained model, characterized by similar learning requirements and adaptation behaviors. 

Motivated by the above observation, we propose a Fisher geometry-based unified task-free framework, named FiUni, for the continual parameter-efficient fine-tuning scenario. The key idea of FiUni is to utilize the same Fisher principal subspace signal to simultaneously perform batch-level latent task detection and LoRA subspace construction. Specifically, FiUni maintains a storehouse of historical Fisher subspaces during online training and matches the Fisher principal subspace of each incoming batch against this storehouse. The resulting subspace similarity serves as a unified geometric signal to adaptively determine whether the current batch should reuse an existing subspace, expand a related subspace, or be assigned to a new subspace. Thus, FiUni jointly models latent task detection and parameter-efficient adaptation within a single framework, enabling the model to dynamically balance knowledge sharing and task-specific isolation without relying on explicit task boundaries or IDs.

The main contributions of this work are summarized as follows: 
\begin{itemize}
    \item We reveal the intrinsic task geometry within Fisher/K-FAC principal subspaces. By utilizing only a small number of downstream samples to estimate the Fisher/K-FAC principal subspace, we depict meaningful task-geometric structures, thereby providing an effective signal for measuring task similarity.
    \item We propose FiUni, a task-free continual parameter-efficient fine-tuning framework that unifies latent task shift detection and LoRA subspace construction via a single geometric signal. Furthermore, we design an adaptive decision mechanism to dynamically balance knowledge sharing and task isolation based on the matching degree between current data and historical Fisher subspaces.
    \item Experimental results demonstrate that FiUni can effectively infer the latent task affiliation of each incoming batch without relying on task boundaries or IDs, and achieve competitive performance compared to state-of-the-art task-aware continual learning baselines while requiring fewer trainable parameters.
\end{itemize}
  

\section{Related Work}
CL \cite{cl_survey1, cl_survey2} aims to enable models to sequentially obtain knowledge from continuous data streams or tasks while preserving performance on previously learned information. The primary challenge in this paradigm is catastrophic forgetting, i.e., adapting to new tasks tends to overwrite or disrupt parameter representations crucial for prior knowledge \cite{cf1, cf2}. To mitigate this issue, existing research broadly categorizes CL approaches into regularization-based, replay-based, and dynamic architecture-based methods. Regularization-based techniques \cite{ewc, si, mas} typically preserve crucial parameters of previous tasks by restricting updates on new tasks. Replay-based strategies \cite{gem, dgr} alleviate forgetting by saving a small number of old task samples or generating historical task samples during the training procedure of new tasks. Moreover, dynamic architecture-based schemes \cite{den, learn_to_grow} progressively expand the network topology to provide additional capacity for new knowledge. Recently, the development of LLMs has shifted the research focus of CL from traditional lightweight classifiers to massive pre-trained models \cite{cl_survey3, cl_survey4}. 



This shift makes parameter efficiency a central concern. PEFT \cite{peft_survey1, peft_survey2} provides a feasible approach by freezing most pre-trained parameters and introducing only a small number of trainable parameters for downstream adaptation. Typical approaches include Adapters \cite{adapter}, Prefix Tuning \cite{prefix-tuning}, and LoRA \cite{lora}. LoRA introduces low-rank update branches into weight matrices, and these updates can be merged back after training without additional inference latency. Building on this low-rank formulation, recent works have further attempted to impose pre-defined update subspaces on LoRA to reduce trainable parameters while maintaining adaptation performance \cite{vera, lora-xs, fourier_lora}.
For instance, FiLoRA \cite{filora} utilizes the K-FAC \cite{k-fac} approximation of the FIM from a small number of downstream samples, to extract principal subspaces as fixed low-rank bases, thereby guiding LoRA to perform efficient adaptation in parameter-sensitive directions. This indicates that the Fisher principal subspace can characterize the important update directions of the current task-specific parameters and provide a more geometrically meaningful subspace selection for efficient parameter fine-tuning.

In CL scenarios, PEFT provides a natural way to manage task-specific adaptation parameters, which can be independently stored, composed, or isolated. 
Building upon this property, existing continual PEFT methods typically enforce orthogonal constraints to separate the low-rank update spaces of different tasks, thereby reducing inter-task interference \cite{o_lora, pecl}.
Furthermore, recent methods have moved beyond strict isolation and explicitly model shared and task-specific components across tasks to improve transfer ability \cite{sparta, ella}.
Although current continual PEFT schemes demonstrate strong performance in task-aware settings, their reliance on explicit task boundaries or task IDs for module management limits their applicability to task-free online scenarios. While recent task-label-free methods have explored model-internal signals to reduce the dependence on task identities \cite{migu}, they mainly focus on update modulation rather than explicitly coordinating latent task discovery with parameter-efficient subspace organization. Therefore, how to unify task-free latent structure identification and adaptive PEFT subspace management remains underexplored.


The intrinsic difference between FiUni and the above methods lies in its ability to determine the allocation of incoming knowledge at the batch level within a task-free setting. 
Furthermore, rather than passively mitigating inter-task interference through auxiliary regularization terms during the training process, FiUni actively confines parameter updates to low-rank subspaces determined by Fisher geometry during the subspace construction stage. 
Importantly, unlike methods that depend on explicit task boundaries to dictate knowledge sharing and isolation strategies, the mechanisms of knowledge reuse and task isolation in FiUni emerge naturally from the Fisher geometry induced by the pre-trained model on a minimal set of downstream samples, rather than being explicitly specified by human-annotated task partitions.

\begin{figure*}[t]
    \centering
    \includegraphics[width=\textwidth]{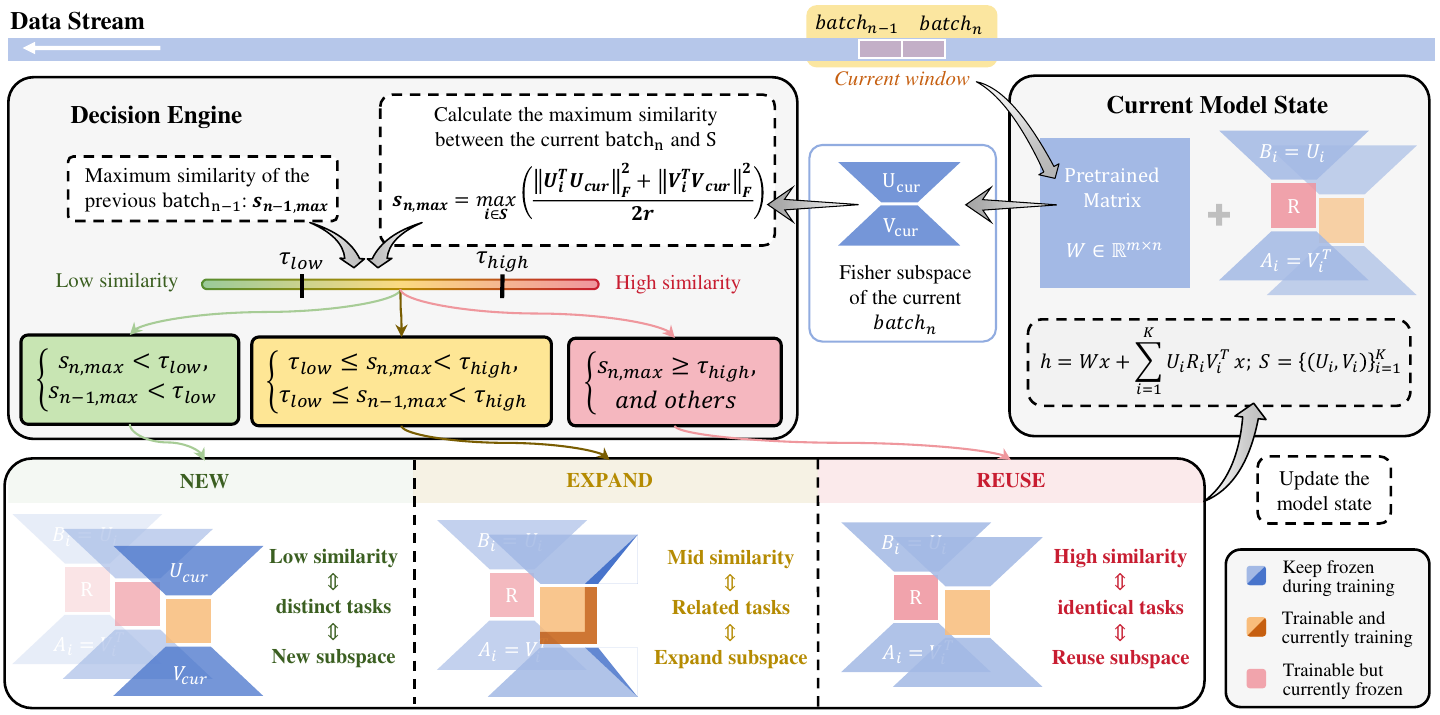}
    \caption{Overview of FiUni. For each incoming batch in a task-free stream, FiUni estimates its Fisher principal subspace and matches it with the historical subspace pool. Based on the subspace similarity, FiUni adaptively decides to create a New subspace, Expand a related subspace, or Reuse an existing one.
    }
    \label{fig2}
\end{figure*}

\section{Methodology}
\subsection{Fisher Subspace Similarity}
\label{subsec:fisher_similarity}
We first introduce how FiUni measures the geometric similarity between different downstream data batches through Fisher principal subspaces. Consider a linear layer in a pre-trained model with weight matrix $W \in \mathbb{R}^{m \times n}$. For a data window $\mathcal{B}$, the Fisher information matrix of this layer can be approximated by K-FAC \cite{k-fac} as the Kronecker product of two covariance factors: $F_W \approx \mathcal{G} \otimes \mathcal{A}$, where
\begin{equation}
    \mathcal{A} = \mathbb{E}_{x \sim \mathcal{B}}[x x^\top],
    \mathcal{G} = \mathbb{E}_{x \sim \mathcal{B}}[\delta \delta^\top].
\end{equation}

Here, $x$ denotes the input activation of the layer, and $\delta$ denotes the back-propagated gradient with respect to the layer output. The activation covariance factor $\mathcal{A}$ captures the dominant directions in the input feature space, while the gradient covariance factor $\mathcal{G}$ captures the dominant sensitive directions in the output gradient space. Together, they characterize the Fisher geometry induced by the current data on the parameter updates of this layer.

To obtain a compact task representation, we perform eigendecomposition on $\mathcal{A}$ and $\mathcal{G}$, and select the top-$r_{\mathrm{det}}$ eigenvectors corresponding to the largest eigenvalues, where $r_{\mathrm{det}}$ denotes the detection rank used for Fisher subspace matching:
\begin{equation}
    V_{\mathcal{B}} = \operatorname{TopEig}(\mathcal{A}, r_{\mathrm{det}}),
    U_{\mathcal{B}} = \operatorname{TopEig}(\mathcal{G}, r_{\mathrm{det}}),
\end{equation}
where $V_{\mathcal{B}} \in \mathbb{R}^{n \times r_{\mathrm{det}}}$ and $U_{\mathcal{B}} \in \mathbb{R}^{m \times r_{\mathrm{det}}}$. We refer to $(U_{\mathcal{B}}, V_{\mathcal{B}})$ as the Fisher principal subspace of the current data window.

For two data windows $\mathcal{B}_i$ and $\mathcal{B}_j$, we measure their geometric similarity by the overlap between their Fisher principal subspaces. Specifically, the gradient-side and activation-side subspace similarities are defined as
\begin{equation}
    s_U(\mathcal{B}_i,\mathcal{B}_j)
    =
    \frac{1}{r_{\mathrm{det}}}
    \left\|
    U_i^\top U_j
    \right\|_F^2,
\end{equation}
\begin{equation}
    s_V(\mathcal{B}_i,\mathcal{B}_j)
    =
    \frac{1}{r_{\mathrm{det}}}
    \left\|
    V_i^\top V_j
    \right\|_F^2.
\end{equation}

The final Fisher subspace similarity is defined as
\begin{equation}
    s(\mathcal{B}_i,\mathcal{B}_j)
    =
    \frac{
    s_U(\mathcal{B}_i,\mathcal{B}_j)
    +
    s_V(\mathcal{B}_i,\mathcal{B}_j)
    }{2}.
    \label{eq:fisher_similarity}
\end{equation}

For a multi-layer model, we compute the above similarity over multiple selected target layers and modules, and average them as the overall task similarity:
\begin{equation}
    s_{ij}
    =
    \frac{1}{|\mathcal{L}|}
    \sum_{\ell \in \mathcal{L}}
    s^{(\ell)}(\mathcal{B}_i,\mathcal{B}_j),
    \label{eq:multi_layer_similarity}
\end{equation}
where $\mathcal{L}$ denotes the set of layers and modules used for Fisher subspace estimation. This similarity can be interpreted as the average principal-direction overlap between two Fisher geometries. A larger $s_{ij}$ suggests that two data windows may come from the same or related latent task phases, while a smaller value indicates stronger orthogonality and may imply the emergence of a new latent phase. Therefore, this metric can be used to detect latent task changes in an online data stream by measuring task similarity between different data windows.


\subsection{FiUni Framework Design}
\label{subsec:fiuni_framework}

Based on the observation that Fisher principal subspaces reflect task similarity, we propose FiUni, a unified task-free continual learning framework for latent task detection and parameter-efficient adaptation, as shown in Figure~\ref{fig2}. FiUni considers an online data stream where training data arrive sequentially in batches:
\begin{equation}
    \mathcal{D}
    =
    \{\mathcal{B}_1, \mathcal{B}_2, \ldots, \mathcal{B}_n, \ldots\},
\end{equation}
During training, we maintain a historical Fisher subspace pool:
\begin{equation}
    \mathcal{S}
    =
    \{(U_k, V_k, R_k)\}_{k=1}^{K},
\end{equation}
where $K$ denotes the number of discovered latent phases. For the $k$-th phase, $U_k$ and $V_k$ are frozen bases derived from Fisher/K-FAC principal subspaces, and $R_k$ is the corresponding trainable core matrix. For a given layer, FiUni represents the effective model parameter as



\begin{equation}
    W_{\mathrm{eff}}
    =
    W_0
    +
    \sum_{k=1}^{K}
    U_k R_k V_k^\top,
    \label{eq:weff}
\end{equation}
where $W_0$ is the frozen pre-trained weight. Unlike standard LoRA, which directly learns a full low-rank decomposition, FiUni fixes the left and right update bases using Fisher principal subspaces and only trains the compact intermediate matrix $R_k$, thereby restricting parameter updates to Fisher-sensitive directions of the current data.


When a new data window $\mathcal{B}_n$ arrives, FiUni first estimates its Fisher principal subspace by extracting the top-$r$ principal directions $(U_{\mathrm{cur}}, V_{\mathrm{cur}})$ for subsequent LoRA subspace construction.

It then computes the similarity between the current subspace and all historical subspaces using the top-$r_{\mathrm{det}}$ principal directions, with $r_{\mathrm{det}} \leq r$, and takes the maximum value as the matching degree between the current window and historical knowledge:
\begin{equation}
    s_{n,\max}
    =
    \max_{k \in \{1,\ldots,K\}}
    s
    \left(
    (U_{\mathrm{cur}}, V_{\mathrm{cur}}),
    (U_k, V_k)
    \right).
    \label{eq:max_similarity}
\end{equation}

This maximum similarity reflects whether the current data can be explained by existing Fisher subspaces. A larger $s_{n,\max}$ indicates that the current data is highly related to a historical phase, while a smaller value suggests that the current data may correspond to a new latent task phase.

Based on the matching degree between the current window and historical subspaces, FiUni adopts a three-state decision mechanism: REUSE, EXPAND, NEW. This mechanism is controlled by two thresholds, $\tau_{\mathrm{low}}$ and $\tau_{\mathrm{high}}$.

When $s_{n,\max} \geq \tau_{\mathrm{high}}$, FiUni regards the current data as highly matched with an existing phase, and thus reuses the corresponding LoRA subspace by continuing to update its core matrix $R_k$. When $\tau_{\mathrm{low}} \leq s_{n,\max} < \tau_{\mathrm{high}}$, FiUni treats the current data as related to an existing phase with moderate drift. It therefore expands the most similar FiLoRA subspace by removing the overlapping components of $(U_{\mathrm{cur}}, V_{\mathrm{cur}})$, orthogonalizing the remaining directions, and supplementing them with random directions to increase the subspace rank by $\Delta r$. When $s_{n,\max} < \tau_{\mathrm{low}}$, FiUni considers the current data geometrically distinct from historical phases and allocates a new LoRA subspace using the current Fisher principal subspace:
\begin{equation}
    U_{K+1} = U_{\mathrm{cur}}, 
    V_{K+1} = V_{\mathrm{cur}},
\end{equation}
and initializes a new trainable core matrix $R_{K+1}$. The new tuple $(U_{K+1}, V_{K+1}, R_{K+1})$ is then added to the historical subspace pool.


To improve online detection stability and reduce false triggers caused by sample noise, FiUni adopts a two-window confirmation mechanism. For two consecutive batches/windows, FiUni performs the corresponding REUSE or NEW operation only when both satisfy the same decision condition, making online task-shift detection more robust.

\subsection{Geometric Reuse and Isolation in FiUni}
Unlike methods that rely on given task boundaries to actively isolate adaptation modules and predefine shared or isolated parameters, knowledge sharing and task isolation in FiUni both arise from Fisher geometry itself, and parameter updates are restricted to the corresponding geometric subspaces.

Taking the output-gradient Fisher subspace $U$ as an example, for the current historical subspace set $\mathcal{S}$, the common overlapping component among these subspaces corresponds to reusable shared directions:
\begin{equation}
    U_{\mathcal{S}}^{\mathrm{share}}
    =
    \bigcap_{k=1}^{K}
    U_k .
\end{equation}

For any subspace $U_i$ and the remaining subspace pool $U_{\neg i}$, the orthogonal component corresponds to more task-specific isolated directions for the latent task:
\begin{equation}
    U_i^{\mathrm{iso}}
    =
    (I - U_{\neg i}U_{\neg i}^{\top})U_i .
\end{equation}

The overlapping components support knowledge reuse across related phases, while the orthogonal residuals provide isolation for new or drifting knowledge. Therefore, FiUni does not simply switch modules after detecting task boundaries. Instead, it performs geometric matching at the batch level and adaptively balances reuse, expansion, and isolation throughout the online data stream.

In practice, estimating Fisher statistics and performing eigendecomposition for all parameter blocks would introduce considerable computational overhead, while our ablation experiments show that the discriminative ability of Fisher subspace similarity is not very sensitive to the specific choice of layers or modules, but is mainly affected by the number of samples used for estimation and the $r_{\mathrm{det}}$. Therefore, FiUni performs subspace matching only on a small number of selected layers. This enables FiUni to perform efficient online detection with limited overhead, making it practical for continual parameter-efficient fine-tuning of large language models.

\begin{figure*}[t]
    \centering
    \includegraphics[width=\textwidth]{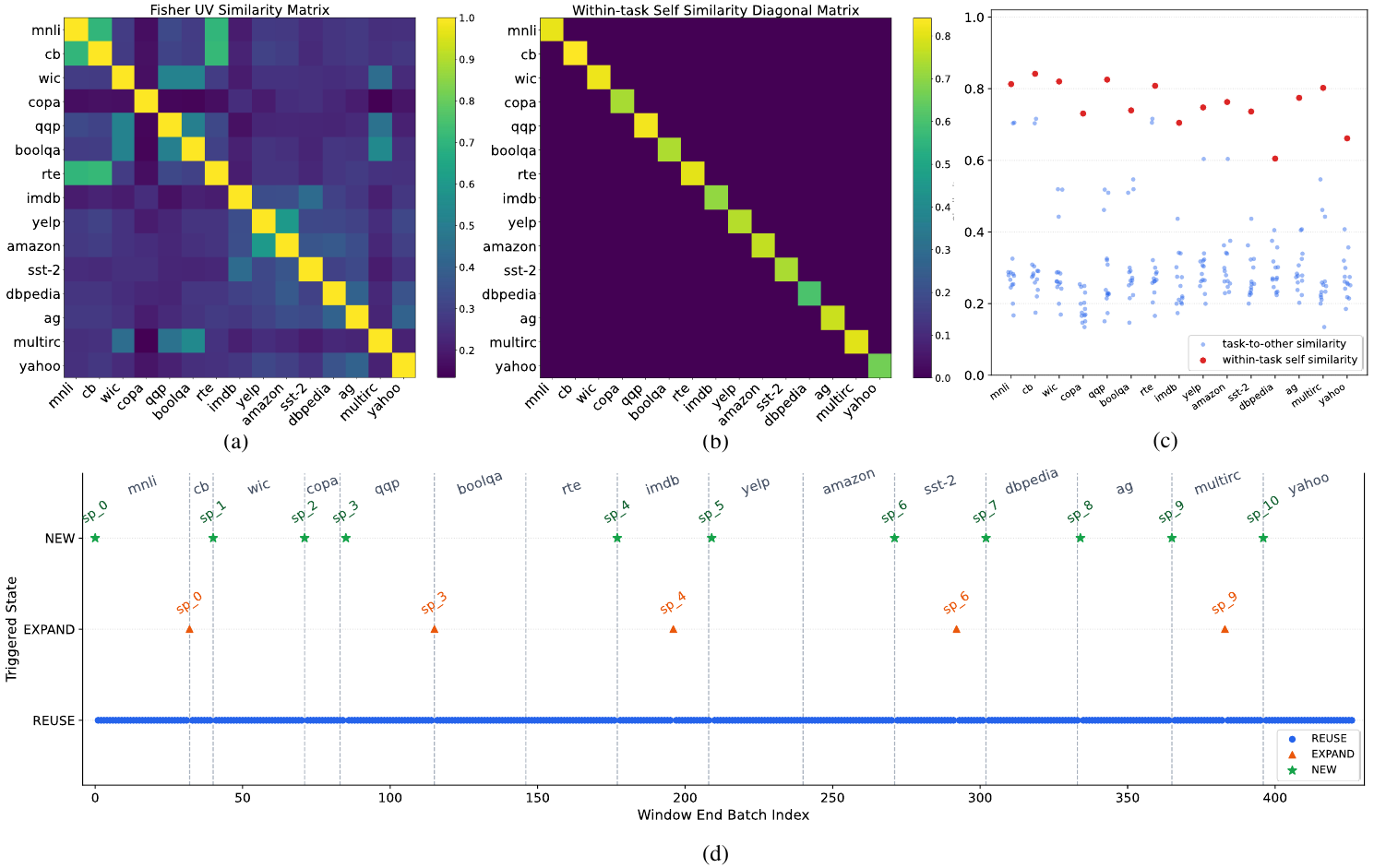}
    \caption{Empirical analysis of Fisher subspace similarity and FiUni decisions on T5-base. Panels (a)--(c) estimate Fisher subspaces using 32 samples per task over all linear layers with $r_{det}=8$: (a) shows cross-task Fisher subspace similarity, (b) shows within-task self-similarity from two independent samplings, and (c) compares within-task similarity with task-to-other similarity. Panel (d) shows the batch-level decisions produced by FiUni on a task-free stream using 32-shot Fisher estimation, selected Q/K/V/O layers, and thresholds $\tau_{\mathrm{low}}=0.65$ and $\tau_{\mathrm{high}}=0.85$.
    }
    \label{fig:fisher_subspace_similarity}
\end{figure*}

\section{Experiment}

\begin{table*}[t]
\centering
\scriptsize
\setlength{\tabcolsep}{3.5pt}
\renewcommand{\arraystretch}{1.05}
\resizebox{\textwidth}{!}{
\begin{tabular}{llccccccccc}
\toprule
\multirow{2}{*}{} & \multirow{2}{*}{\textbf{Methods}} 
& \multicolumn{4}{c}{\textbf{Standard CL Benchmark (SC)}}
& \multicolumn{4}{c}{\textbf{Long Sequence Benchmark (LS)}}
& \multicolumn{1}{c}{\textbf{TRACE}} \\
\cmidrule(lr){3-6} \cmidrule(lr){7-10} \cmidrule(lr){11-11}
& & \textbf{Order 1} & \textbf{Order 2} & \textbf{Order 3} & \textbf{Avg} 
& \textbf{Order 4} & \textbf{Order 5} & \textbf{Order 6} & \textbf{Avg}
& \textbf{Order 7}  \\
\midrule

\multirow{11}{*}{\rotatebox{90}{T5-Large}}
& SeqLoRA\textsuperscript{\dag} & 39.5 & 31.9 & 46.6 & 39.3 & 4.9 & 3.5 & 4.2 & 4.2 & 12.1 \\
& \textcolor{gray}{SeqLoRAReplay} & \textcolor{gray}{74.0} & \textcolor{gray}{73.1} & \textcolor{gray}{73.0} & \textcolor{gray}{73.3} & \textcolor{gray}{74.2} & \textcolor{gray}{72.7} & \textcolor{gray}{73.9} & \textcolor{gray}{73.6} & \textcolor{gray}{34.0} \\
& IncLoRA & 63.4 & 62.2 & 65.1 & 63.6 & 63.0 & 57.9 & 60.4 & 60.5 & - \\
& EWC \cite{ewc} & 46.3 & 45.3 & 52.1 & 47.9 & 44.9 & 44.0 & 45.4 & 44.8 & - \\
& L2P \cite{l2p} & 60.3 & 61.7 & 61.1 & 60.7 & 57.5 & 53.8 & 56.9 & 56.1 & - \\
& \textcolor{gray}{LFPT5 \cite{lfpt5}} & \textcolor{gray}{67.6} & \textcolor{gray}{72.6} & \textcolor{gray}{77.9} & \textcolor{gray}{72.7} & \textcolor{gray}{70.4} & \textcolor{gray}{68.2} & \textcolor{gray}{69.1} & \textcolor{gray}{69.2} & - \\
& O-LoRA \cite{o_lora} & 73.2 & \underline{72.4} & 70.4 & 72.0 & \underline{69.9} & \underline{68.5} & 65.3 & 67.9 & \underline{23.1} \\
& + MIGU \cite{migu} & 73.5 & 71.4 & 70.0 & 71.6 & 65.4 & 65.2 & 65.2 & 65.3 & - \\
& SpaRTA \cite{sparta} & \underline{73.7} & 70.5 & \textbf{73.8} & \underline{72.7} & \textbf{71.5} & \textbf{70.5} & \underline{68.0} & \textbf{70.0} & 16.7 \\
& \textcolor{gray}{+ Replay} & \textcolor{gray}{77.0} & \textcolor{gray}{75.6} & \textcolor{gray}{75.2} & \textcolor{gray}{75.9} & \textcolor{gray}{75.6} & \textcolor{gray}{73.2} & \textcolor{gray}{74.1} & \textcolor{gray}{74.3} & \textcolor{gray}{36.5} \\
& FiUni (ours)\textsuperscript{\dag} & \textbf{75.6} & \textbf{76.5} & \underline{73.0} & \textbf{75.0} & 67.1 & 67.0 & \textbf{70.84} & \underline{68.3} & \textbf{32.9} \\

\midrule 

\multirow{7}{*}{\rotatebox{90}{LLaMA-3.1-8B}}
& SeqLoRA\textsuperscript{\dag} & 75.88 & 74.40 & 74.35 & 74.86 & 67.81 & 65.93 & 63.80 & 65.85 & 28.23 \\
& O-LoRA \cite{o_lora} & 69.39 & 67.58 & 71.44 & 69.46 & 69.54 & 64.42 & 66.50 & 66.82 & 28.45 \\
& SpaRTA \cite{sparta} & 76.10 & 75.69 & 75.52 & 75.77 & 71.34 & 70.77 & 72.95 & 71.69 & 31.33 \\
& \textcolor{gray}{+ Replay} & \textcolor{gray}{77.56} & \textcolor{gray}{77.01} & \textcolor{gray}{76.03} & \textcolor{gray}{76.83} & \textcolor{gray}{73.10} & \textcolor{gray}{73.05} & \textcolor{gray}{74.17} & \textcolor{gray}{73.44} & \textcolor{gray}{34.16} \\
& \textcolor{gray}{SeqLoRAReplay} & \textcolor{gray}{77.02} & \textcolor{gray}{76.53} & \textcolor{gray}{76.19} & \textcolor{gray}{76.58} & \textcolor{gray}{72.03} & \textcolor{gray}{72.96} & \textcolor{gray}{75.38} & \textcolor{gray}{73.46} & \textcolor{gray}{33.02} \\
& ELLA \cite{ella} & \underline{77.80} & \underline{77.20} & \textbf{77.70} & \underline{77.57} & \underline{72.87} & \underline{72.84} & \underline{76.82} & \underline{74.18} & \underline{33.29} \\
& FiUni (ours)\textsuperscript{\dag} & \textbf{78.00} & \textbf{78.20} & \underline{77.49} & \textbf{77.91} & \textbf{75.74} & \textbf{74.74} & \textbf{77.50} & \textbf{75.99} & \textbf{55.21} \\
\bottomrule
\end{tabular}
}
\caption{Comparison results on Standard CL Benchmark (SC), Long Sequence Benchmark (LS), and TRACE. FiUni results are averaged over three random seeds. Methods marked with \textsuperscript{\dag} do not rely on task boundaries during training. Gray-colored methods use replay data, generated pseudo-data, or external memory during continual learning. \textbf{Bold} indicates the best result, and \underline{underline} indicates the second-best result among non-replay methods.}
\label{tab:main_results}
\end{table*}

\subsection{Benchmarks and Data Setup}
\label{subsec:benchmarks}
We evaluate FiUni on three CL benchmarks. Unlike conventional task-aware continual learning, we focus on the more challenging task-free online setting, where training data arrive sequentially as a batch stream and the model has no access to explicit task boundaries or task IDs during training.

\textbf{Standard CL Benchmark (SC)} consists of four text classification datasets: AG News, Amazon Reviews, DBpedia, and Yahoo Answers \cite{sc_benchmark}. Following prior work \cite{o_lora}, we organize them into three task orders, denoted as Orders 1--3, to evaluate continual adaptation over short classification sequences.

\textbf{Long Sequence Benchmark (LS)} extends SC to a longer and more diverse stream with 15 tasks \cite{ls_benchmark}. We sample a fixed number of training examples for each task and construct three task orders, denoted as Orders 4--6. Compared with SC, LS contains more heterogeneous task types and more complex task transitions.

\textbf{TRACE Benchmark} \cite{trace} contains diverse LLM tasks, including multiple-choice QA, multilingual understanding, code generation, and mathematical reasoning, denoted as Order 7.

\subsection{Task Similarity}

We first examine the core observation of this work: Fisher subspace similarity \(s\), estimated from a small number of downstream samples, can reflect geometric similarity between tasks. Specifically, we use T5-base on the LS benchmark and sample 32 examples from each task to compute the pairwise Fisher subspace similarity \(s\) defined in Section~\ref{subsec:fisher_similarity}. We then analyze whether the resulting similarity structure is consistent with task semantics or task families.

As shown in Figure~\ref{fig:fisher_subspace_similarity} (a), the Fisher subspace similarity matrix exhibits clear task structure. Data from identical or related tasks usually show higher subspace similarity, while unrelated tasks tend to be more orthogonal. For example, MNLI, CB, and RTE, which are all related to language inference, show relatively high similarity. Among sentiment-related tasks, IMDB, SST-2, Yelp, and Amazon are all sentiment-oriented, but Yelp and Amazon exhibit stronger similarity, likely because both are rating-style review tasks.

We further evaluate the within-task consistency of this signal through repeated sampling, as shown in Figure~\ref{fig:fisher_subspace_similarity} (b). For each task, we independently sample two groups of examples and estimate their Fisher principal subspaces separately. The results show that two subsets from the same task usually produce highly similar Fisher subspaces, indicating that samples within the same task tend to share consistent Fisher-geometric characteristics. Finally, Figure~\ref{fig:fisher_subspace_similarity} (c) compares within-task self-similarity and task-to-other similarity. Most tasks exhibit substantially higher self-similarity than cross-task similarity. In particular, similarities between identical or related tasks are mostly concentrated in the range of 0.6--0.9, whereas low-similarity or unrelated task pairs mostly fall below 0.6. This suggests that appropriate thresholds can separate similar tasks from unrelated ones, supporting the feasibility of Fisher subspace similarity as a batch-level signal for latent task identification.

\subsection{Task Detection}
After verifying that Fisher subspaces reflect task similarity, we further analyze the batch-level decision process of FiUni in an online task-free stream. We take up to 1,000 samples from each task in the LS benchmark and concatenate them into a sequential training stream. During detection, the model has no access to task IDs or task boundaries, and makes decisions batch by batch.

Figure~\ref{fig:fisher_subspace_similarity} (d) shows the batch-level decision trajectory of FiUni in the task-free stream. When the stream remains within the same task or moves to a related task, FiUni usually tends to reuse an existing subspace or perform a lightweight expansion. When the stream enters a more distinct stage, FiUni can trigger the allocation of a new subspace. For example, within the SST-2 stage, FiUni expands the current subspace when the internal data variation becomes larger. In contrast, when the stream transitions from CB to WiC or from WiC to COPA, FiUni creates a new subspace.

We also observe that FiUni's decisions do not always coincide with human-defined task boundaries. For some related tasks, FiUni may reuse or expand an existing subspace instead of immediately creating a new one. For instance, when the stream moves from MNLI to CB, FiUni expands the subspace associated with MNLI rather than allocating a new subspace. During the Amazon stage, FiUni directly reuses or continues training on the subspace previously triggered by Yelp. This is consistent with our definition of latent tasks. Such behavior allows FiUni to avoid mechanically allocating a separate subspace for every dataset-level task, reducing redundant parameter growth while promoting knowledge reuse across related tasks.

\subsection{Overall Results}
\label{subsec:overall_results}



To verify the effectiveness of FiUni in detecting latent tasks, we compare it with multiple task-aware continual learning methods. Since task boundaries are unavailable in our stream setting, we only consider overall accuracy \cite{oa} $\mathrm{OA}_{T} = \frac{1}{T}\sum_{t=1}^{T} a_{T,t}$, where $a_{T,t}$ denotes the performance on task $t$ after training on the complete stream. We do not consider FWT \cite{gem} or BWT \cite{bwt}, as these metrics rely on task boundaries to measure forgetting and knowledge transfer.



As shown in Table~\ref{tab:main_results}, FiUni achieves strong results across different benchmarks and backbone models, and outperforms strict orthogonal isolation methods such as O-LoRA in all settings. This demonstrates the effectiveness of using fixed Fisher subspaces for task sharing and isolation. We observe that FiUni brings relatively limited improvement on LS with T5-Large. A possible reason is that T5-Large has a lower hidden dimension, and to keep the FiLoRA parameter scale competitive with standard LoRA, the available rank capacity can be consumed earlier in the long task stream, thereby restricting the expansion space for later phases.

In contrast, LLaMA-3.1-8B provides a larger representation space and richer subspace capacity, where FiUni achieves more consistent gains across the three benchmarks. Overall, without using task boundaries or task IDs, FiUni still matches or surpasses state-of-the-art task-aware continual PEFT methods. This shows that latent task differences can be effectively captured by the Fisher geometry of downstream data on the pre-trained model, enabling effective continual adaptation even without explicit task information.


\section{Conclusion}
We propose FiUni, a Fisher geometry-based framework for task-free continual parameter-efficient fine-tuning in online data streams. FiUni unifies batch-level latent task identification and LoRA subspace construction through Fisher principal subspaces, enabling adaptive reuse, expansion, and allocation of LoRA subspaces without explicit task boundaries or task IDs. Experiments demonstrate that Fisher subspace similarity captures meaningful and stable task geometry, and that FiUni achieves competitive performance against strong task-aware baselines with substantially fewer trainable parameters. Our results highlight Fisher geometry as an effective and practical signal for scalable task-free continual adaptation of LLMs.

\section{Limitations}

FiUni still has several limitations. First, the detection stage estimates Fisher/K-FAC statistics on the pre-trained model, which introduces an extra forward and backward pass. Although this cost is moderate because detection uses only a few samples and selected layers, it may still be noticeable in large-scale online training. A promising future direction is to further study the relationship between task streams and the Fisher factors estimated on the pre-trained model combined with already learned LoRA modules, so that detection can potentially be performed with fewer additional passes or reused statistics. Second, computing the activation and gradient covariance factors also introduces additional memory overhead, since activations and gradients need to be cached for selected modules. Although FiUni avoids constructing the full Fisher matrix, more memory-efficient estimation strategies, such as streaming covariance updates, low-precision statistics, or randomized eigendecomposition, could further improve scalability. Finally, due to computational and memory constraints, our experiments are limited to models up to 8B parameters. We therefore have not fully assessed the behavior and overhead of FiUni on larger-scale models such as 70B LLMs. Further scaling experiments would be needed to verify its efficiency and robustness in such settings.

\section{Ethical Considerations}
FiUni is a method for parameter-efficient continual adaptation and does not introduce new data collection, human annotation, or user-facing decision procedures. All experiments are conducted on publicly available benchmarks following standard research protocols. Since FiUni does not require storing previous task samples for replay, it does not add extra data-retention requirements beyond the training data used in each benchmark. As with standard fine-tuning methods for large language models, practical deployment should follow the data usage policies and safety requirements of the target application.
\section{AI writing statement}
This paper utilized AI assistance for language polishing of the manuscript, including vocabulary correction and spell checking.

\bibliography{ref}
\newpage
\appendix
\section*{Appendix}
This appendix provides additional details and analyses to support the main paper. 
We organize the appendix into three parts. 
First, we provide a theoretical justification for why Fisher information can be used to characterize task-dependent geometry, and further explain why the K-FAC principal subspaces of the activation and gradient covariance factors can reflect task similarity. 
Second, we present detailed experimental settings, including benchmark construction, implementation details, hyperparameters, and evaluation protocols. 
Third, we report additional experimental results and ablation studies to further analyze the behavior of FiUni under different configurations.

\section{Theoretical Justification}
\label{app:theory}

In this section, we provide a theoretical justification for why Fisher information can be used to identify different latent tasks, and why its K-FAC approximation allows us to use the principal subspaces of the activation and gradient covariance factors as a practical task-similarity signal. We first discuss the full Fisher information matrix, and then derive the connection under the K-FAC approximation.

\subsection{Full Fisher Information as Task Geometry}

Consider a pre-trained model with parameters $\theta_0$ and a downstream task distribution $\mathcal{D}_t$. For a sample $z=(x,y)\sim \mathcal{D}_t$, let $\ell(z;\theta_0)$ denote the training loss, and define the task-induced gradient as
\begin{equation}
    g_t(z)
    =
    \nabla_{\theta}
    \ell(z;\theta)
    \big|_{\theta=\theta_0}.
\end{equation}
The Fisher information matrix \cite{natural-gradient} induced by task $\mathcal{D}_t$ is
\begin{equation}
    F_t
    =
    \mathbb{E}_{z\sim \mathcal{D}_t}
    \left[
    g_t(z)g_t(z)^\top
    \right].
\end{equation}

This matrix captures the second-order geometry of the loss around the pre-trained model. For a small parameter perturbation $\Delta \theta$, the local change of the task loss or predictive distribution can be approximated by a quadratic form:
\begin{equation}
    \Delta \mathcal{L}_t
    \approx
    \frac{1}{2}
    \Delta \theta^\top
    F_t
    \Delta \theta.
\end{equation}
Therefore, directions with large Fisher eigenvalues correspond to parameter directions to which task $\mathcal{D}_t$ is most sensitive. In other words, the dominant eigenspace of $F_t$ describes the main update directions required by this task on the pre-trained model.

Let
\begin{equation}
    Q_t
    =
    \operatorname{TopEig}(F_t,r)
\end{equation}
be the subspace spanned by the top-$r$ eigenvectors of $F_t$. For two tasks $\mathcal{D}_a$ and $\mathcal{D}_b$, their Fisher principal subspaces $Q_a$ and $Q_b$ reflect whether the two tasks require similar parameter update directions. A natural similarity measure is
\begin{equation}
    s_F(a,b)
    =
    \frac{1}{r}
    \left\|
    Q_a^\top Q_b
    \right\|_F^2.
\end{equation}
If the two tasks induce similar gradient distributions on the pre-trained model, then their Fisher matrices are close, and their principal subspaces are also close. If the two tasks require very different update directions, their Fisher principal subspaces tend to have smaller overlap.

This can be formalized using standard eigenspace perturbation results. Suppose the top-$r$ eigenspace of $F_a$ has eigengap $\gamma_a$, and $F_b$ is a perturbation of $F_a$. Then the distance between their principal subspaces satisfies
\begin{equation}
    \left\|
    \sin\Theta(Q_a,Q_b)
    \right\|_2
    \leq
    C
    \frac{
    \left\|
    F_a-F_b
    \right\|_2
    }{
    \gamma_a
    },
\end{equation}
where $C$ is a constant and $\Theta(Q_a,Q_b)$ denotes the principal angles between the two subspaces. This indicates that if two tasks induce similar Fisher matrices, their principal subspaces remain aligned; if their Fisher matrices differ significantly, their principal subspaces become less aligned or more orthogonal.

Thus, the full Fisher information matrix provides a task-dependent geometric representation: its leading eigenspace captures the parameter-sensitive directions of a task, and the overlap between two Fisher principal subspaces reflects the similarity between the corresponding tasks.

\subsection{K-FAC Approximation of the Layer-wise Fisher Matrix}

Although the full Fisher information matrix provides a principled task geometry, it is prohibitively expensive to compute and store for large language models. Therefore, we use the K-FAC \cite{k-fac} approximation to obtain a tractable layer-wise representation.

Consider a linear layer with weight matrix
\begin{equation}
    W \in \mathbb{R}^{m\times n}.
\end{equation}
Let $x\in \mathbb{R}^{n}$ be the input activation of this layer, and let $\delta\in \mathbb{R}^{m}$ be the back-propagated gradient with respect to the layer output. The gradient of the loss with respect to $W$ is
\begin{equation}
    \nabla_W \ell
    =
    \delta x^\top.
\end{equation}
After vectorization, this gradient can be written as
\begin{equation}
    \operatorname{vec}(\nabla_W \ell)
    =
    x \otimes \delta.
\end{equation}
The layer-wise Fisher matrix is then
\begin{equation}
    F_W
    =
    \mathbb{E}
    \left[
    \operatorname{vec}(\nabla_W \ell)
    \operatorname{vec}(\nabla_W \ell)^\top
    \right].
\end{equation}

K-FAC approximates this Fisher matrix by factorizing the second-order statistics of activations and output gradients:
\begin{equation}
    F_W
    \approx
    \mathcal{G}
    \otimes
    \mathcal{A},
\end{equation}
where
\begin{equation}
\begin{aligned}
    \mathcal{A}
    &=
    \mathbb{E}
    \left[
    x x^\top
    \right],
    \\
    \mathcal{G}
    &=
    \mathbb{E}
    \left[
    \delta \delta^\top
    \right].
\end{aligned}
\end{equation}
Here, $\mathcal{A}$ captures the dominant directions in the input activation space, while $\mathcal{G}$ captures the dominant directions in the output-gradient space. Together, they approximate the layer-wise Fisher geometry induced by the current data.

\subsection{Why the Principal Subspaces of \texorpdfstring{$\mathcal{A}$}{A} and \texorpdfstring{$\mathcal{G}$}{G} Reflect Task Similarity}

We now explain why the top eigenvectors of $\mathcal{A}$ and $\mathcal{G}$ can be used to reflect task similarity under the K-FAC approximation.

Let
\begin{equation}
\begin{aligned}
    \mathcal{A}v_i
    &=
    \alpha_i v_i,
    \\
    \mathcal{G}u_j
    &=
    \beta_j u_j.
\end{aligned}
\end{equation}
Then the Kronecker product $\mathcal{G}\otimes \mathcal{A}$ has eigenvectors of the form $u_j\otimes v_i$, with eigenvalues $\beta_j\alpha_i$:
\begin{equation}
    (\mathcal{G}\otimes \mathcal{A})
    (u_j\otimes v_i)
    =
    \beta_j\alpha_i
    (u_j\otimes v_i).
\end{equation}
Therefore, the dominant eigenspace of the K-FAC Fisher matrix is determined by the dominant eigenspaces of $\mathcal{G}$ and $\mathcal{A}$. This means that the leading eigenvectors of the gradient covariance factor and the activation covariance factor provide a compact approximation to the principal geometry of the layer-wise Fisher matrix.

For a data window or task $\mathcal{B}$, we define
\begin{equation}
\begin{aligned}
    U_{\mathcal{B}}
    &=
    \operatorname{TopEig}(\mathcal{G}_{\mathcal{B}}, r),
    \\
    V_{\mathcal{B}}
    &=
    \operatorname{TopEig}(\mathcal{A}_{\mathcal{B}}, r).
\end{aligned}
\end{equation}
Here, $U_{\mathcal{B}}$ represents the dominant output-gradient directions, and $V_{\mathcal{B}}$ represents the dominant input-activation directions. Since the dominant directions of $\mathcal{G}\otimes \mathcal{A}$ are constructed from combinations of $U_{\mathcal{B}}$ and $V_{\mathcal{B}}$, the pair $(U_{\mathcal{B}},V_{\mathcal{B}})$ can serve as a tractable representation of the Fisher principal geometry.

Now consider two data distributions or tasks $\mathcal{D}_a$ and $\mathcal{D}_b$. If they are similar, they tend to produce similar activation patterns and similar output-gradient patterns on the same pre-trained model. This implies that
\begin{equation}
    \mathcal{A}_a
    \approx
    \mathcal{A}_b,
\end{equation}
and
\begin{equation}
    \mathcal{G}_a
    \approx
    \mathcal{G}_b.
\end{equation}
Under eigengap conditions, the top eigenspaces of these matrices will also be close. Thus, $V_a$ will be close to $V_b$, and $U_a$ will be close to $U_b$. Conversely, if two tasks induce very different activation or gradient covariance structures, their top eigenspaces will have smaller overlap.

This provides the theoretical basis for using the principal subspaces of $\mathcal{A}$ and $\mathcal{G}$ to compare tasks. Instead of explicitly constructing the full Fisher eigenspace, FiUni measures the overlap between the two K-FAC factors:
\begin{equation}
    s_U(\mathcal{B}_i,\mathcal{B}_j)
    =
    \frac{1}{r_{\mathrm{det}}}
    \left\|
    U_i^\top U_j
    \right\|_F^2,
\end{equation}
\begin{equation}
    s_V(\mathcal{B}_i,\mathcal{B}_j)
    =
    \frac{1}{r_{\mathrm{det}}}
    \left\|
    V_i^\top V_j
    \right\|_F^2.
\end{equation}
The final similarity is
\begin{equation}
    s(\mathcal{B}_i,\mathcal{B}_j)
    =
    \frac{
    s_U(\mathcal{B}_i,\mathcal{B}_j)
    +
    s_V(\mathcal{B}_i,\mathcal{B}_j)
    }{2}.
\end{equation}

This similarity is a surrogate for the overlap between the dominant eigenspaces of the K-FAC Fisher matrices. A high value means that the two data windows share similar activation and gradient principal directions, and therefore induce similar Fisher geometry. A low value means that their sensitive update directions are more orthogonal, suggesting that they correspond to different latent tasks or learning phases.

\subsection{Implication for Task-free Continual Learning}

The above analysis shows that Fisher information contains task-dependent geometry at two levels. At the full-matrix level, the Fisher principal subspace captures the dominant parameter directions to which a task is most sensitive. Under the K-FAC approximation, this geometry is represented by the principal subspaces of the activation covariance factor $\mathcal{A}$ and the gradient covariance factor $\mathcal{G}$.

This explains why the Fisher/K-FAC principal subspaces can be used for latent task identification in an online task-free stream. If an incoming batch belongs to an existing latent task, its Fisher principal subspace should have high similarity with historical subspaces. If it corresponds to a new or substantially different latent phase, its Fisher subspace should have low similarity to all historical subspaces. Intermediate similarity naturally corresponds to partial overlap, motivating the expansion operation in FiUni.

Therefore, Fisher geometry provides a unified signal for both detecting the latent task affiliation of each incoming batch and constructing the corresponding low-rank adaptation subspace.

\paragraph{Connection to subspace Methods.}
The above analysis also provides a possible explanation for why subspace methods, such as SeqLoRA and O-LoRA, can be effective in continual learning. We conjecture that, around a pre-trained model, different downstream tasks do not explore the entire parameter space uniformly, but instead tend to update along Fisher-sensitive directions induced by their task data. As a result, the effective update space used by each task may occupy only a low-dimensional subspace of the full parameter space. When different tasks correspond to distinct Fisher-sensitive directions, stochastic gradient descent may naturally push their updates toward partially separated or nearly orthogonal regions during training.

From this perspective, orthogonal constraints and subspace allocation mechanisms may not create task isolation from scratch; rather, they may reinforce the implicit geometric separation induced jointly by the pre-trained model and downstream tasks. Thus, the effectiveness of these methods may not solely come from the imposed orthogonality constraints, but may also rely on a task-related geometric structure that already exists near the pre-trained model: different tasks correspond to different dominant Fisher directions, while related tasks share partially overlapping directions. This perspective suggests that parameter isolation and knowledge reuse in continual learning can be understood as explicit ways of exploiting this pre-training-induced geometric structure.

\section{System-level Explanation of FiUni}
\label{app:system_explanation}


Notably, the Fisher subspace similarity based detection mechanism is not tied to a specific LoRA adaptation design. Since it only relies on the geometric matching between the Fisher principal subspace of the current batch and historical subspaces, it can serve as a standalone upstream module for latent task identification and be plugged into other continual learning methods or task-expert selection scenarios. FiUni further combines this detection mechanism with Fisher-guided LoRA subspace construction to enable task-free continual parameter-efficient fine-tuning.

\section{Experimental Settings}
\label{app:experiment}

\begin{table*}[t]
\centering
\small
\setlength{\tabcolsep}{5.5pt}
\renewcommand{\arraystretch}{1.08}
\begin{tabular}{l l l l l}
\toprule
\textbf{Dataset Name} & \textbf{Category} & \textbf{Task} & \textbf{Domain} & \textbf{Metric} \\
\midrule
Yelp     & CL Benchmark & Sentiment Analysis & Yelp Reviews & Accuracy \\
Amazon   & CL Benchmark & Sentiment Analysis & Amazon Reviews & Accuracy \\
DBpedia  & CL Benchmark & Topic Classification & Wikipedia & Accuracy \\
Yahoo    & CL Benchmark & Topic Classification & Yahoo Q\&A & Accuracy \\
AG News  & CL Benchmark & Topic Classification & News & Accuracy \\
MNLI     & GLUE & Natural Language Inference & Various & Accuracy \\
QQP      & GLUE & Paraphrase Detection & Quora & Accuracy \\
RTE      & GLUE & Natural Language Inference & News, Wikipedia & Accuracy \\
SST-2    & GLUE & Sentiment Analysis & Movie Reviews & Accuracy \\
WiC      & SuperGLUE & Word Sense Disambiguation & Lexical Databases & Accuracy \\
CB       & SuperGLUE & Natural Language Inference & Various & Accuracy \\
COPA     & SuperGLUE & Question Answering & Blogs, Encyclopedia & Accuracy \\
BoolQA   & SuperGLUE & Boolean Question Answering & Wikipedia & Accuracy \\
MultiRC  & SuperGLUE & Question Answering & Various & Accuracy \\
IMDB     & SuperGLUE & Sentiment Analysis & Movie Reviews & Accuracy \\
\bottomrule
\end{tabular}
\caption{
Task details of the Long Sequence Benchmark (LS). The first five datasets correspond to the standard CL benchmark, while the remaining tasks are drawn from GLUE, SuperGLUE, and IMDB.
}
\label{tab:ls_tasks}
\end{table*}

\begin{table*}[t]
\centering
\small
\setlength{\tabcolsep}{7pt}
\renewcommand{\arraystretch}{1.08}
\begin{tabular}{l l c l l r}
\toprule
\textbf{Dataset} & \textbf{Source} & \textbf{Avg. Len.} & \textbf{Metric} & \textbf{Language} & \textbf{\#Data} \\
\midrule
\rowcolor{gray!18}
\multicolumn{6}{l}{\textit{Domain-specific}} \\
ScienceQA   & Science & 210  & Accuracy & English & 5,000 \\
FOMC        & Finance & 51   & Accuracy & English & 5,000 \\
MeetingBank & Meeting & 2853 & ROUGE-L  & English & 5,000 \\
\midrule
\rowcolor{gray!18}
\multicolumn{6}{l}{\textit{Multi-lingual}} \\
C-STANCE  & Social Media & 127 & Accuracy & Chinese & 5,000 \\
20Minuten & News & 382 & SARI & German & 5,000 \\
\midrule
\rowcolor{gray!18}
\multicolumn{6}{l}{\textit{Code Completion}} \\
Py150 & GitHub & 422 & Edit Similarity & Python & 5,000 \\
\midrule
\rowcolor{gray!18}
\multicolumn{6}{l}{\textit{Mathematical Reasoning}} \\
NumGLUE-cm & Math & 32 & Accuracy & English & 5,000 \\
NumGLUE-ds & Math & 21 & Accuracy & English & 5,000 \\
\bottomrule
\end{tabular}
\caption{
Task details of TRACE. The benchmark covers domain-specific tasks, multilingual tasks, code completion, and mathematical reasoning.
}
\label{tab:trace_tasks}
\end{table*}

\begin{table*}[t]
\centering
\small
\setlength{\tabcolsep}{6pt}
\renewcommand{\arraystretch}{1.32}
\begin{tabular}{M{3.5cm} M{0.8cm} L{10cm}}
\toprule
\textbf{Benchmark} & \textbf{Order} & \textbf{Task Sequence} \\
\midrule
\multirow{3}{=}{\centering Standard CL Benchmark}
& 1 & DBpedia $\rightarrow$ Amazon $\rightarrow$ Yahoo $\rightarrow$ AG News \\[2pt]
& 2 & DBpedia $\rightarrow$ Amazon $\rightarrow$ AG News $\rightarrow$ Yahoo \\[2pt]
& 3 & Yahoo $\rightarrow$ Amazon $\rightarrow$ AG News $\rightarrow$ DBpedia \\
\midrule
\multirow{5.25}{=}{\centering Long Sequence Benchmark}
& 4 & MNLI $\rightarrow$ CB $\rightarrow$ WiC $\rightarrow$ COPA $\rightarrow$ QQP $\rightarrow$ BoolQA $\rightarrow$ RTE $\rightarrow$ IMDB $\rightarrow$ Yelp $\rightarrow$ Amazon $\rightarrow$ SST-2 $\rightarrow$ DBpedia $\rightarrow$ AG News $\rightarrow$ MultiRC $\rightarrow$ Yahoo \\[4pt]

& 5 & MultiRC $\rightarrow$ BoolQA $\rightarrow$ WiC $\rightarrow$ MNLI $\rightarrow$ CB $\rightarrow$ COPA $\rightarrow$ QQP $\rightarrow$ RTE $\rightarrow$ IMDB $\rightarrow$ SST-2 $\rightarrow$ DBpedia $\rightarrow$ AG News $\rightarrow$ Yelp $\rightarrow$ Amazon $\rightarrow$ Yahoo \\[4pt]

& 6 & Yelp $\rightarrow$ Amazon $\rightarrow$ MNLI $\rightarrow$ CB $\rightarrow$ COPA $\rightarrow$ QQP $\rightarrow$ RTE $\rightarrow$ IMDB $\rightarrow$ SST-2 $\rightarrow$ DBpedia $\rightarrow$ AG News $\rightarrow$ Yahoo $\rightarrow$ MultiRC $\rightarrow$ BoolQA $\rightarrow$ WiC \\
\midrule
\multirow{1.75}{=}{\centering TRACE}
& 7 & C-STANCE $\rightarrow$ FOMC $\rightarrow$ MeetingBank $\rightarrow$ Py150 $\rightarrow$ ScienceQA $\rightarrow$ NumGLUE-cm $\rightarrow$ NumGLUE-ds $\rightarrow$ 20Minuten \\
\bottomrule
\end{tabular}
\caption{
Task orders used in our continual learning experiments. Orders 1--3 correspond to SC, Orders 4--6 correspond to LS, and Order 7 corresponds to TRACE.
}
\label{tab:task_orders}
\end{table*}

\subsection{Benchmarks}
\label{app:benchmarks}

We evaluate FiUni on three continual learning benchmarks: \textbf{Standard CL Benchmark (SC)} \cite{sc_benchmark}, \textbf{Long Sequence Benchmark (LS)} \cite{ls_benchmark}, and \textbf{TRACE} \cite{trace}. These benchmarks cover task streams of different lengths, domains, and task formats. SC focuses on short text classification sequences, LS extends the setting to longer and more heterogeneous classification-oriented task streams by incorporating additional tasks from GLUE \cite{glue} and SuperGLUE \cite{superglue}, and TRACE further introduces diverse LLM-oriented tasks such as multilingual understanding, code completion, mathematical reasoning, and question answering. The benchmark composition and task orders are summarized in Tables~\ref{tab:ls_tasks}, \ref{tab:trace_tasks}, and \ref{tab:task_orders}.

For the SC and LS benchmarks, we follow the same data sampling protocol as prior studies \cite{o_lora, sparta}. Specifically, we use the same task orders and the same sampled training and evaluation splits provided in their official code repositories, ensuring that both training and testing data are consistent with existing baselines for fair comparison.

In addition to hiding task boundaries, our setting differs from task-aware continual learning in how the data stream is constructed. 
We concatenate datasets from different tasks into a single continuous stream. 
As a result, around a manually defined task boundary, a training batch may contain samples from two substantially different datasets. 
This makes our task-free benchmark closer to realistic online scenarios and more challenging than the standard task-aware setting.

\subsection{Baselines}
\label{app:baselines}

We compare FiUni with a broad set of continual learning and parameter-efficient fine-tuning baselines. These baselines cover sequential LoRA training, replay-based methods, regularization-based continual learning, prompt-based continual learning, and recent orthogonal or subspace-based continual PEFT methods.

\noindent\textbf{SeqLoRA.}
SeqLoRA sequentially trains LoRA modules on the task stream without preserving previous task knowledge or allocating task specific subspaces. It serves as a simple continual PEFT baseline that directly applies LoRA to sequential learning.

\noindent\textbf{SeqLoRAReplay.}
SeqLoRAReplay extends SeqLoRA with experience replay. During training on new tasks, it reuses a small number of stored examples from previous tasks to reduce forgetting.

\noindent\textbf{IncLoRA.}
IncLoRA incrementally introduces new LoRA parameters for newly encountered tasks. Instead of using a single shared LoRA module, it expands adaptation parameters along the task sequence to provide additional capacity for new knowledge.

\noindent\textbf{EWC} \cite{ewc}.
Elastic Weight Consolidation is a regularization based continual learning method. It estimates parameter importance, typically using Fisher information, and penalizes changes to parameters important for previous tasks.

\noindent\textbf{L2P} \cite{l2p}.
Learning to Prompt maintains a pool of learnable prompts and selects input relevant prompts during training. It adapts to new tasks by reusing and updating prompt knowledge from previous tasks.

\noindent\textbf{LFPT5} \cite{lfpt5}.
LFPT5 continuously trains soft prompts for T5 based continual learning. The learned prompts are used both for task solving and pseudo sample generation, which supports experience replay.

\noindent\textbf{O-LoRA} \cite{o_lora}.
O-LoRA reduces task interference by enforcing orthogonality among the low rank update spaces of different tasks. In this way, task specific parameter updates are isolated in separate subspaces.

\noindent\textbf{MIGU} \cite{migu}.
MIGU is a rehearsal free and task label free continual learning method. It exploits the magnitude distribution of outputs in linear layers and updates parameters with large output magnitudes to reduce gradient conflicts across tasks.

\noindent\textbf{SpaRTA} \cite{sparta}.
SpaRTA is a continual PEFT method that manages task specific and shared adaptation components. It aims to reduce forgetting while preserving transfer across related tasks.

\noindent\textbf{ELLA} \cite{ella}.
ELLA is a continual PEFT baseline that models task relationships in the adaptation space. It encourages knowledge sharing across related tasks while maintaining task specific components to reduce interference.

\subsection{Implementation Details}
\label{app:implementation_details}

\begin{table*}[t]
\centering
\small
\setlength{\tabcolsep}{5pt}
\renewcommand{\arraystretch}{1.15}
\begin{tabular}{l C{1.35cm} C{1.35cm} C{1.35cm} C{1.35cm} C{1.35cm} C{1.35cm}}
\toprule
\multirow{2}{*}{\textbf{Hyperparameter}} 
& \multicolumn{3}{c}{\textbf{T5-large}} 
& \multicolumn{3}{c}{\textbf{Llama-3.1-8B}} \\
\cmidrule(lr){2-4}
\cmidrule(lr){5-7}
& \textbf{SC} & \textbf{LS} & \textbf{TRACE} 
& \textbf{SC} & \textbf{LS} & \textbf{TRACE} \\
\midrule
Batch size 
& \multicolumn{6}{c}{32} \\

Learning rate 
& \multicolumn{3}{c}{1e-3} 
& \multicolumn{3}{c}{1e-4} \\

Epochs 
& \multicolumn{6}{c}{1} \\

FiLoRA rank $r$ 
& \multicolumn{3}{c}{32} 
& \multicolumn{3}{c}{128} \\

Dropout 
& 0 & 0.05 & 0 & 0.1 & 0.1 & 0 \\

Adaptation layers $\mathcal{L}_{\mathrm{adapt}}$ 
& \multicolumn{6}{c}{all} \\

Target modules $\mathcal{M}$ 
& \multicolumn{3}{c}{q,k,v,o} 
& \multicolumn{3}{c}{q,v} \\
\bottomrule
\end{tabular}
\caption{
Training and adaptation hyperparameters for FiUni. 
Batch size denotes the effective batch size, computed as the raw batch size multiplied by the number of gradient accumulation steps.
}
\label{tab:adapt_hyperparameters}
\end{table*}

\begin{table*}[t]
\centering
\small
\setlength{\tabcolsep}{5pt}
\renewcommand{\arraystretch}{1.15}
\begin{tabular}{l C{1.35cm} C{1.35cm} C{1.35cm} C{1.35cm} C{1.35cm} C{1.35cm}}
\toprule
\multirow{2}{*}{\textbf{Hyperparameter}} 
& \multicolumn{3}{c}{\textbf{T5-large}} 
& \multicolumn{3}{c}{\textbf{Llama-3.1-8B}} \\
\cmidrule(lr){2-4}
\cmidrule(lr){5-7}
& \textbf{SC} & \textbf{LS} & \textbf{TRACE} 
& \textbf{SC} & \textbf{LS} & \textbf{TRACE} \\
\midrule
Detection rank $r_{\mathrm{det}}$ 
& \multicolumn{3}{c}{4} 
& \multicolumn{3}{c}{8} \\

Expansion rank $\Delta r$ 
& 0 & 4 & 4 & 8 & 8 & 8 \\

$\tau_{\mathrm{low}}$ 
& \multicolumn{3}{c}{0.5} 
& \multicolumn{3}{c}{0.6} \\

$\tau_{\mathrm{high}}$ 
& \multicolumn{6}{c}{0.7} \\

Detection layers $\mathcal{L}_{\mathrm{det}}$ 
& \multicolumn{3}{c}{23} 
& \multicolumn{3}{c}{31} \\

Expand cooldown $C_{\mathrm{expand}}$ 
& 50 & 200 & 100 & 40 & 40 & 40 \\

New cooldown $C_{\mathrm{new}}$ 
& 30 & 10 & 100 & 10 & 10 & 10 \\
\bottomrule
\end{tabular}
\caption{
Detection and decision hyperparameters for FiUni.
}
\label{tab:detect_hyperparameters}
\end{table*}

We implement all experiments based on Transformers 4.57.6 \cite{transformers} and PyTorch 2.6.0 \cite{pytorch}. 
To ensure consistency across different methods, all baselines and FiUni are evaluated under the same software environment and hardware configuration. 
All experiments are conducted on a single NVIDIA RTX 6000 Ada Generation GPU. 
All models are initialized in bfloat16 (bf16) precision to reduce memory usage and computational overhead. 
Unless otherwise specified, all methods use the same backbone model, task orders, data splits, and evaluation protocol within each benchmark to ensure fair comparison.

For the SC and LS benchmarks, we concatenate all training data according to the predefined task orders and feed them into the model as a continuous batch stream. 
This setting simulates an online task-free continual learning scenario, where data arrive sequentially and the model must update itself based only on the current batch. 
During training, the model has no access to task boundaries or task IDs, and therefore cannot rely on explicit task-switching signals to create, select, or reset adaptation modules.

For the TRACE benchmark, different tasks involve different batch size requirements. 
Therefore, for data-format compatibility, we organize training batches task by task rather than directly mixing all samples into a single stream. 
However, this does not provide task-aware supervision to the model: task identities and explicit task-switching signals are still hidden during training. 
For each incoming batch, its latent task affiliation and the corresponding subspace management decision are determined adaptively by FiUni based on Fisher subspace similarity.

We summarize the hyperparameters used for different backbones and benchmarks in Tables~\ref{tab:adapt_hyperparameters} and~\ref{tab:detect_hyperparameters}. 
Here, $r$ denotes the LoRA/FiLoRA subspace rank, $r_{\mathrm{det}}$ denotes the detection rank used for Fisher subspace matching, and $\Delta r$ denotes the additional rank introduced during expansion. 
The thresholds $\tau_{\mathrm{low}}$ and $\tau_{\mathrm{high}}$ control the New, Expand, and Reuse decisions. 
We use $\mathcal{L}_{\mathrm{adapt}}$ for the layers used in adaptation, $\mathcal{L}_{\mathrm{det}}$ for the layers used in Fisher-based detection, and $\mathcal{M}$ for the target modules. 
The cooldown steps $C_{\mathrm{expand}}$ and $C_{\mathrm{new}}$ prevent Expand and New operations from being triggered too frequently.

\section{Additional Results}
\label{app:additional_results}

\subsection{Ablation on Fisher Subspace Similarity Computation}
\label{app:similarity_ablation}

We further analyze how the computation of Fisher subspace similarity $s$ is affected by different design choices, including the detection rank $r_{\mathrm{det}}$, the number of samples used for Fisher estimation, the selected modules, and the selected layers. 
Unless otherwise specified, the ablation is conducted on the LS benchmark using the same similarity computation protocol as in the main paper. 
Each figure reports both the task-level Fisher subspace similarity matrix and the scatter plot comparing within-task self-similarity with task-to-other similarity.

\begin{figure*}[t]
    \centering
    \includegraphics[width=\textwidth]{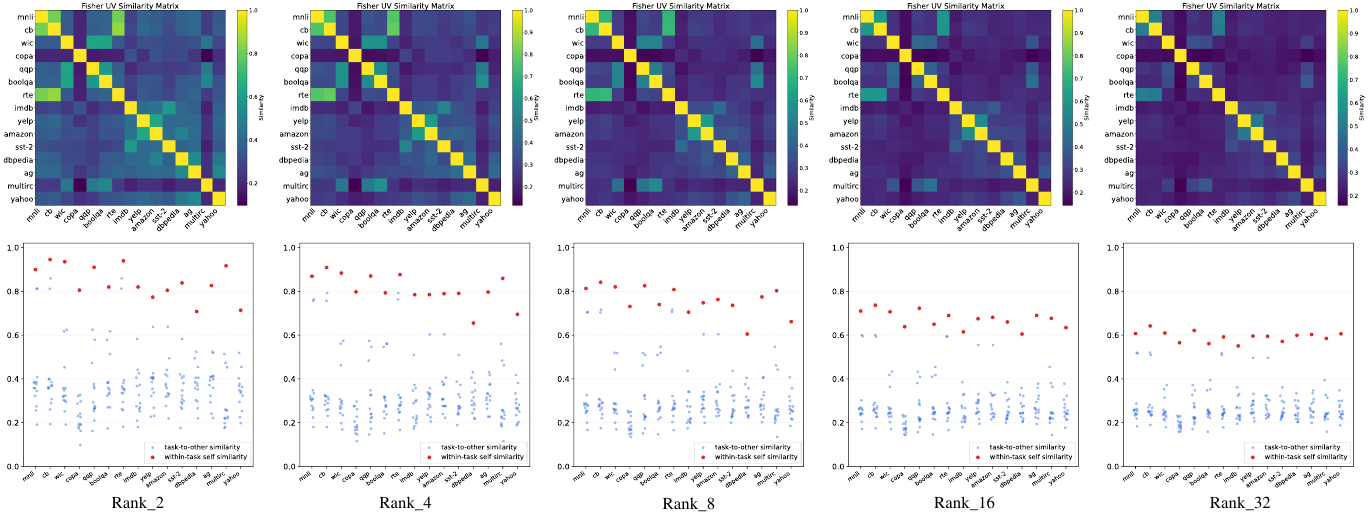}
    \caption{
    Ablation on the detection rank $r_{\mathrm{det}}$ for Fisher subspace similarity computation.
    A moderate rank provides a stable separation between within-task self-similarity and task-to-other similarity.
    }
    \label{fig:ablation_rank}
\end{figure*}

\paragraph{Effect of detection rank.}
Figure~\ref{fig:ablation_rank} shows the effect of using different detection ranks. 
When $r_{\mathrm{det}}$ is too small, the estimated Fisher subspace may not contain enough task-discriminative directions, leading to weaker separation between self-similarity and task-to-other similarity. 
As $r_{\mathrm{det}}$ increases, the similarity structure becomes more stable and better reflects task-level relationships. 
However, using an overly large rank may also introduce less informative directions, making the similarity signal less concentrated. 
This motivates the use of a moderate detection rank in FiUni.

\begin{figure*}[t]
    \centering
    \includegraphics[width=\textwidth]{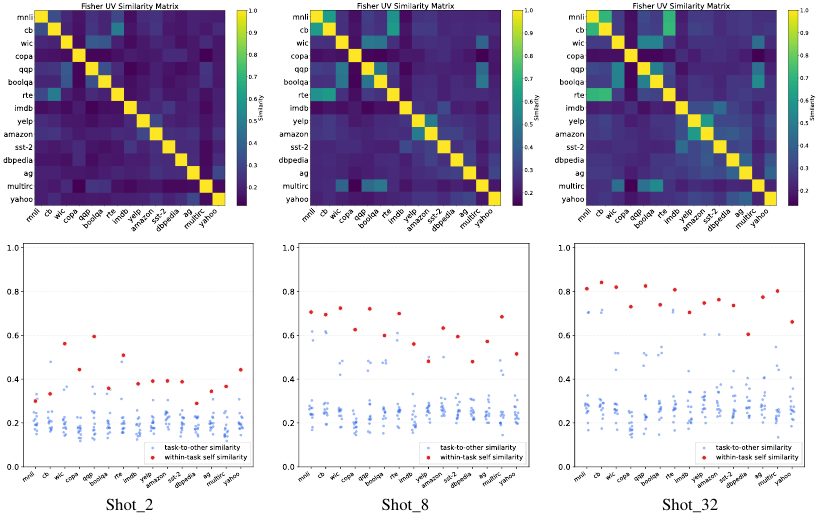}
    \caption{
    Ablation on the number of samples used for Fisher estimation.
    More samples lead to more stable Fisher subspaces and clearer separation between within-task and cross-task similarities.
    }
    \label{fig:ablation_shot}
\end{figure*}

\begin{figure*}[t]
    \centering
    \includegraphics[width=\textwidth]{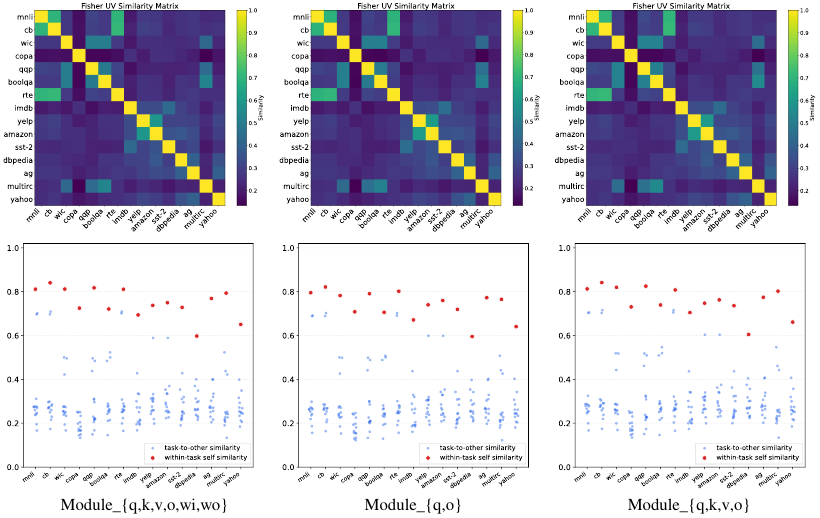}
    \caption{
    Ablation on module selection for Fisher subspace similarity computation.
    Different module choices preserve similar task-level structures, showing that the similarity signal is not tied to a single module configuration.
    }
    \label{fig:ablation_module}
\end{figure*}

\begin{figure*}[t]
    \centering
    \includegraphics[width=\textwidth]{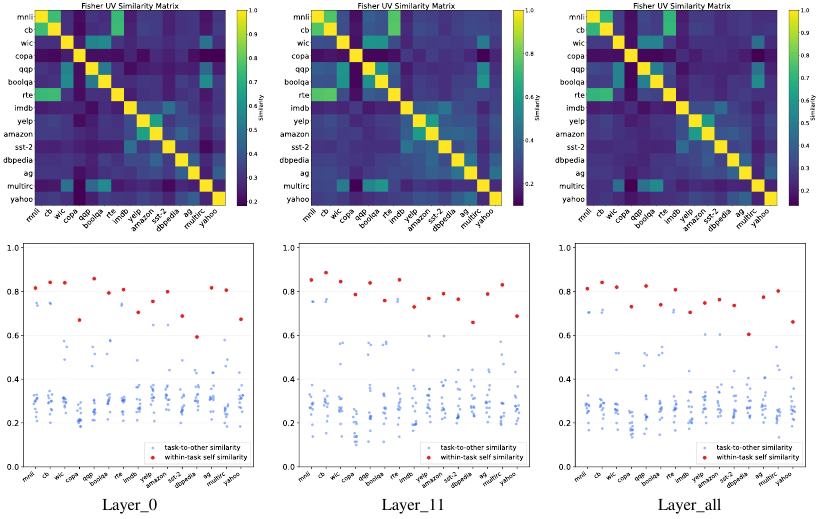}
    \caption{
    Ablation on layer selection for Fisher subspace similarity computation.
    The resulting similarity structures remain broadly consistent across different layer choices, supporting efficient detection with selected layers.
    }
    \label{fig:ablation_layer}
\end{figure*}
\paragraph{Effect of Fisher estimation samples.}
Figure~\ref{fig:ablation_shot} studies the influence of the number of samples used for estimating the Fisher/K-FAC factors. 
With very few samples, the estimated Fisher statistics are noisier, and the resulting similarity signal becomes less stable. 
Increasing the number of samples improves the consistency of within-task self-similarity and makes the separation from unrelated tasks clearer. 
These results show that Fisher subspace similarity can be estimated from a small number of downstream samples, while a sufficient sample size is still helpful for robust task discrimination.

\paragraph{Effect of module selection.}
Figure~\ref{fig:ablation_module} compares different choices of Transformer modules for Fisher subspace estimation. 
The results indicate that the similarity pattern is relatively robust across different module selections. 
In particular, attention-related modules already provide meaningful task-discriminative signals, suggesting that FiUni does not require estimating Fisher statistics on all modules to obtain useful task geometry. 
This supports our design of performing detection only on selected modules to reduce computation.

\paragraph{Effect of layer selection.}
Figure~\ref{fig:ablation_layer} evaluates whether Fisher subspace similarity depends strongly on the choice of layers. 
We observe that different layer selections produce broadly consistent task-similarity patterns. 
This suggests that the task-discriminative Fisher geometry is not limited to a specific layer, and reliable detection can be achieved using only a subset of layers. 
Therefore, FiUni can reduce online detection overhead by computing Fisher subspace similarity on selected layers rather than all parameter blocks.

Overall, these ablations show that Fisher subspace similarity is a robust signal for capturing task-level geometry. 
Although the quality of the signal is affected by the number of estimation samples and the detection rank, it remains relatively stable across different module and layer selections. 
This supports FiUni's practical design of using a moderate detection rank, a small number of Fisher estimation samples, and only selected modules/layers for efficient online task-free detection.
Meanwhile, although the overall similarity values decrease under extremely few-shot estimation and the separation between related and unrelated tasks becomes less pronounced, the within-task similarity still remains the highest for most tasks.
This suggests that Fisher subspace similarity can still preserve a certain degree of task-discriminative signal in low-sample regimes, providing a practical basis for sample-efficient batch-level detection in downstream continual adaptation.

\subsection{Training Dynamics and Decision Trajectory}
\label{app:loss_decision}

\begin{figure*}[t]
    \centering
    \includegraphics[width=\textwidth]{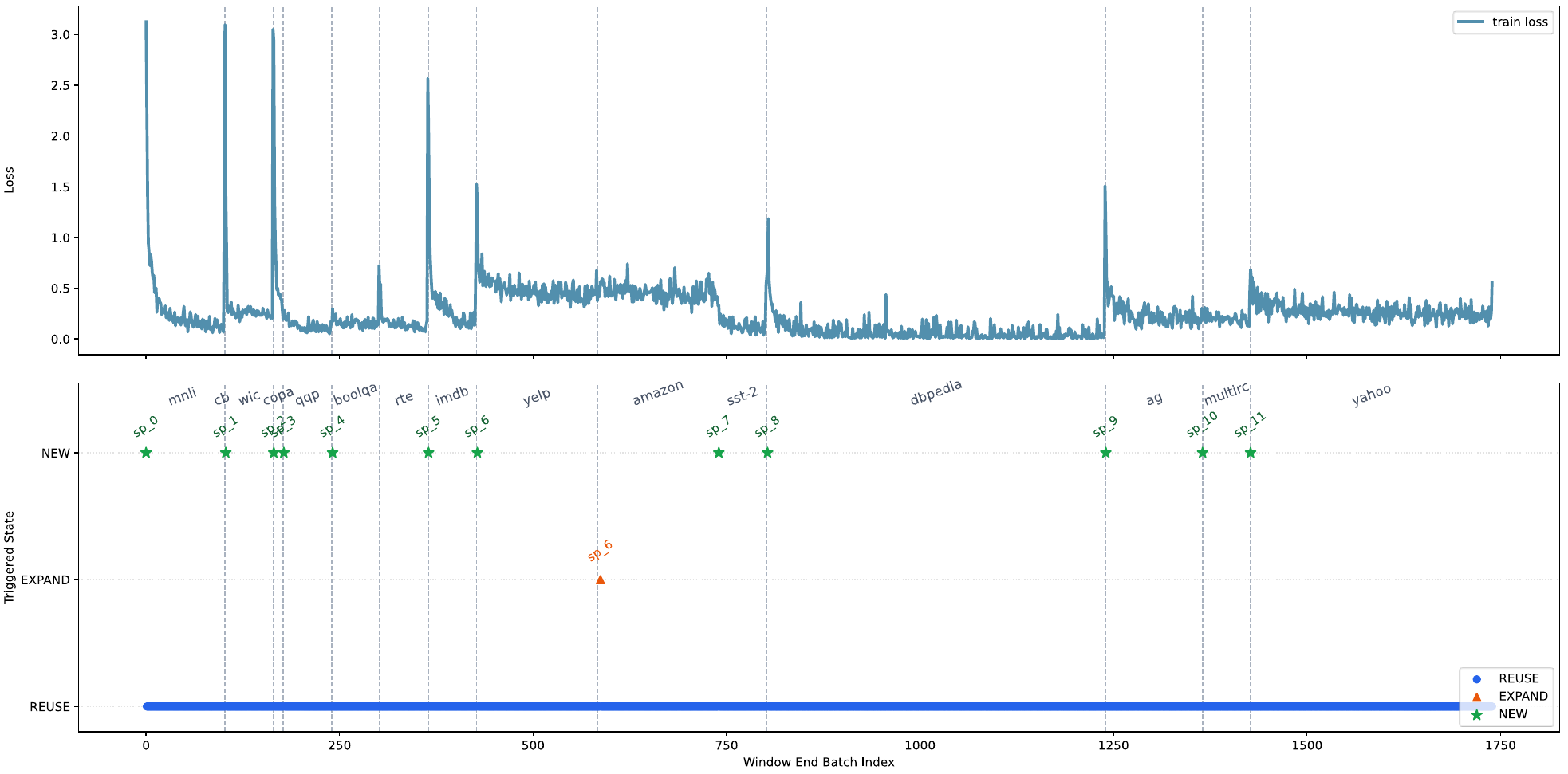}
    \caption{
    Training loss and FiUni decision trajectory on the LS benchmark with T5-large. 
    The upper panel shows the training loss along the online data stream, and the lower panel shows the corresponding batch-level \textbf{Reuse}, \textbf{Expand}, and \textbf{New} decisions. 
    }
    \label{fig:loss_decision_t5_ls}
\end{figure*}

\begin{figure*}[t]
    \centering
    \includegraphics[width=\textwidth]{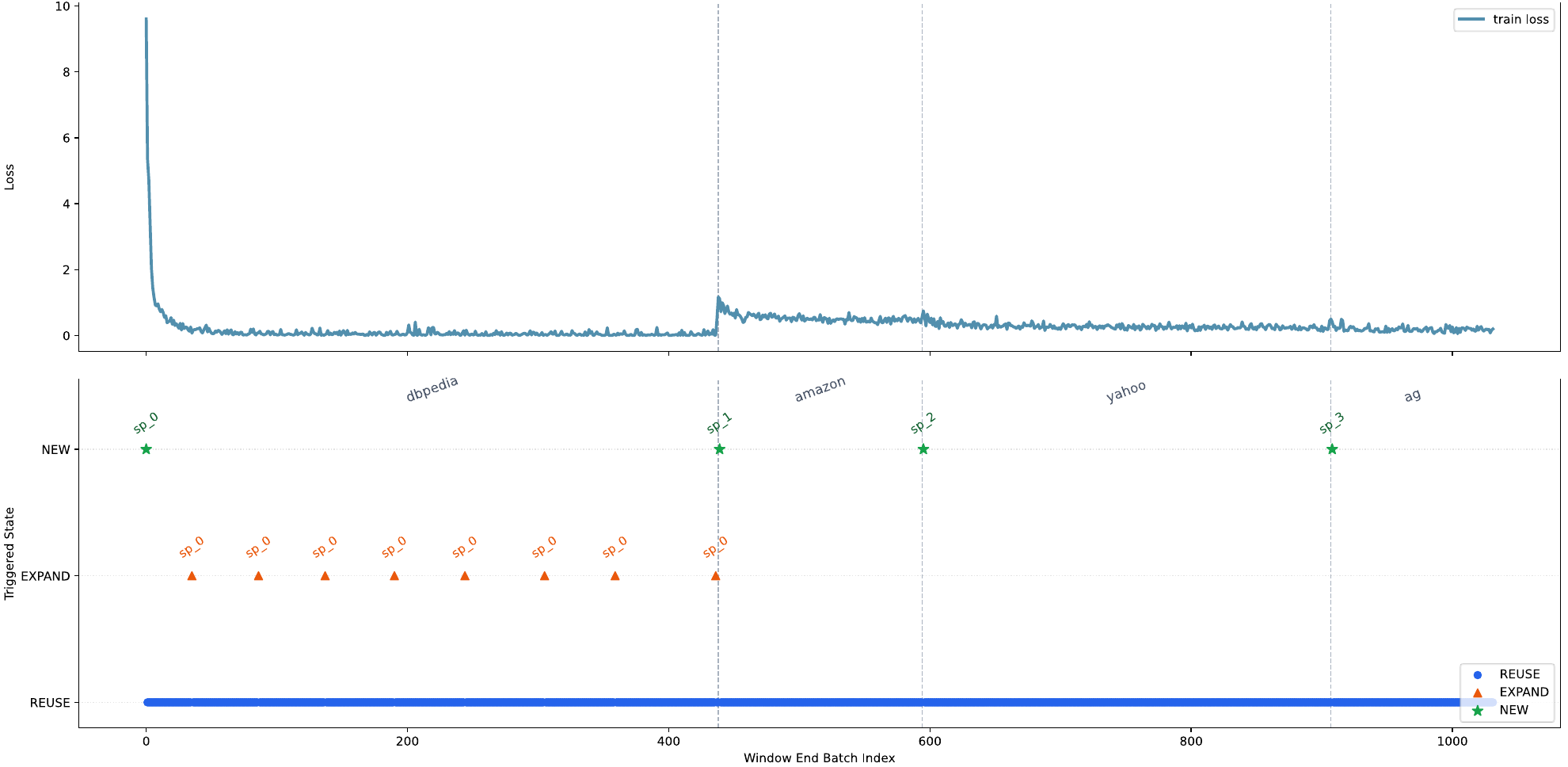}
    \caption{
    Training loss and FiUni decision trajectory on the SC benchmark with T5-large. 
    Even when some task transitions do not produce obvious loss spikes, FiUni can still identify the corresponding latent shifts through Fisher subspace matching and trigger appropriate batch-level decisions.
    }
    \label{fig:loss_decision_t5_sc}
\end{figure*}

We further analyze the online behavior of FiUni by jointly examining the training loss and the corresponding batch-level decision trajectory. Figures~\ref{fig:loss_decision_t5_ls} and~\ref{fig:loss_decision_t5_sc} present the results of T5-large on the LS and SC benchmarks, respectively. In each figure, the upper panel shows the training loss along the sequential data stream, and the lower panel shows the corresponding Reuse, Expand, and New decisions.

Across both benchmarks, the decision trajectory follows the training dynamics in a meaningful yet nontrivial manner. When the incoming data become substantially less compatible with previously accumulated knowledge, the loss often rises and FiUni correspondingly activates New or Expand to introduce additional subspace capacity. By contrast, when the stream remains relatively stable, the decisions are dominated by Reuse, meaning that the current batches can be effectively absorbed by existing Fisher subspaces. Such a pattern is particularly visible on the LS benchmark, where several major shifts in the stream are accompanied by noticeable changes in the loss curve and the appearance of newly allocated subspaces.

At the same time, FiUni is not merely reacting to abrupt loss changes. Compared with methods that rely on sudden loss variation for task-boundary detection, FiUni operates at a finer batch-level granularity through Fisher subspace matching. Consequently, its decision process is able to capture latent task transitions even when the loss curve does not exhibit a pronounced spike. For example, in the SC benchmark, the transitions from Amazon to Yahoo and from Yahoo to AG do not lead to particularly sharp loss peaks, yet FiUni still introduces new subspaces at the corresponding phases. A similar phenomenon appears in the LS benchmark, such as the transitions from Amazon to SST-2 and from AG to MultiRC, where the loss changes only mildly while the decision trajectory still detects the underlying shift in task geometry.

Another noteworthy characteristic is that FiUni does not simply replicate manually defined task boundaries. Some transitions are handled through Reuse or Expand rather than immediately opening a new subspace, especially when adjacent tasks remain geometrically related. 
Likewise, within a single human-defined task, FiUni may occasionally expand an existing subspace when the incoming batches reveal additional internal variation. This behavior is consistent with our notion of latent tasks: the model tracks its own perceived learning phases rather than mechanically following dataset-level segmentation.

In addition, in the SC benchmark, we observe several Expand decisions within the DBpedia stage. Although DBpedia is treated as a single dataset-level task, it covers a wide range of entity categories and topic types, which can induce noticeable batch-level variation in Fisher geometry. 
These batches may not be fully explained by the initial DBpedia subspace, but they remain related enough to avoid opening entirely new subspaces. FiUni therefore expands the existing subspace to provide additional capacity for intra-task variation while preserving knowledge reuse. The different behavior of DBpedia in SC and LS further reflects the context-dependent nature of FiUni's decisions. Rather than assigning a fixed decision pattern to each dataset, FiUni compares every incoming batch with the current historical Fisher subspace pool. In SC, DBpedia appears when the historical pool is still small, making its intra-task variation more likely to fall into the intermediate-similarity region and trigger Expand. In LS, however, DBpedia is encountered after many previous tasks have already populated the subspace pool, so some DBpedia batches can be better explained by existing related subspaces and are thus handled through Reuse. This behavior further supports that FiUni tracks model-perceived latent learning phases rather than simply following dataset names or manually defined task boundaries.

Taken together, the loss-decision analysis provides an intuitive view of how FiUni manages subspaces throughout online continual adaptation. 
Large shifts in the stream are often accompanied by subspace creation or expansion, while stable periods are mainly handled through knowledge reuse. 
More importantly, the decision mechanism is not limited to coarse signals derived from loss fluctuations, but can also resolve finer latent transitions at the batch level, which is especially valuable in realistic task-free settings where explicit task boundaries are unavailable.

\subsection{Dynamic Trainable Parameter Analysis}
\label{app:dynamic_params}

We further analyze the number of trainable parameters from a dynamic perspective on the LS benchmark with T5-large. Different from task-aware methods that typically allocate or optimize a fixed amount of adaptation parameters for each task, FiUni adjusts its trainable parameter budget according to the online decision trajectory. Specifically, a Reuse decision does not introduce additional subspace parameters, while Expand only increases the capacity of an existing subspace by a small amount. A New decision allocates a new LoRA subspace only when the incoming batch is geometrically distinct from all historical subspaces. Therefore, the trainable parameter count of FiUni grows adaptively with the data stream rather than linearly following the number of manually defined tasks.

For comparison, according to the hyperparameter settings reported in prior work \cite{o_lora, sparta}, we compute the accumulated trainable parameters of O-LoRA and SpaRTA on T5-large, while for FiUni we directly count the dynamically allocated trainable parameters after training on Order 4. The comparison is summarized in Table~\ref{tab:trainable_params_order4}.

\begin{table}[t]
\centering
\small
\setlength{\tabcolsep}{6pt}
\renewcommand{\arraystretch}{1.12}
\resizebox{\columnwidth}{!}{
\begin{tabular}{l c c c}
\toprule
\textbf{Metric} & \textbf{O-LoRA} & \textbf{SpaRTA} & \textbf{FiUni (ours)} \\
\midrule
Trainable Params & 35.39M & 44.24M & 3.62M \\
\bottomrule
\end{tabular}
}
\caption{
Trainable parameter comparison on the LS benchmark. 
For FiUni, we report the dynamic trainable parameter count obtained on Order 4.
}
\label{tab:trainable_params_order4}
\end{table}

The results show that FiUni uses substantially fewer trainable parameters than the fixed task-level allocation adopted by strong task-aware baselines. Since FiUni only trains the core matrix within each Fisher-guided LoRA subspace, the trainable parameter count is substantially reduced at the subspace level. On top of this, the adaptive mechanism further helps control parameter growth: related batches can share existing subspaces, moderately shifted batches only expand the most relevant subspace, and new subspaces are created only when necessary. As a result, FiUni avoids assigning a full set of adaptation parameters to every dataset-level task, while still preserving enough flexibility to handle distributional changes in the online stream. This adaptive parameter allocation is especially beneficial for long-sequence continual learning, where blindly adding a complete adapter for each task would lead to rapid parameter growth.



\end{document}